\documentclass[conference]{IEEEtran}
\IEEEoverridecommandlockouts

\usepackage{cite}
\usepackage{amsmath,amssymb,amsfonts}
\usepackage{algorithmic}
\usepackage{graphicx}
\usepackage{textcomp}
\usepackage{xcolor}
\usepackage{url} 
\usepackage{multirow}
\def\BibTeX{{\rm B\kern-.05em{\sc i\kern-.025em b}\kern-.08em
    T\kern-.1667em\lower.7ex\hbox{E}\kern-.125emX}}
\begin{document}

\title{NEVLP: Noise-Robust Framework for Efficient Vision-Language Pre-training}


\author{\IEEEauthorblockN{Yiyi Tao\IEEEauthorrefmark{1}, Zhuoyue Wang\IEEEauthorrefmark{2}, Hang Zhang\IEEEauthorrefmark{3} and Lun Wang\IEEEauthorrefmark{4}}
\IEEEauthorblockA{\IEEEauthorrefmark{1}Johns Hopkins University, MD, USA}
\IEEEauthorblockA{\IEEEauthorrefmark{2}University of California, Berkeley, CA, USA} \IEEEauthorblockA{\IEEEauthorrefmark{3}University of California San Diego, CA, USA}
\IEEEauthorblockA{\IEEEauthorrefmark{4}Duke University, NC, USA}}

\maketitle

\begin{abstract}
The success of Vision Language Models (VLMs) on various vision-language tasks heavily relies on pre-training with large scale web-crawled datasets. However, the noisy and incomplete nature of web data makes dataset scale crucial for performance, rendering end-to-end training increasingly prohibitive. In this paper, we propose NEVLP, a noise-robust framework for efficient vision-language pre-training that requires less pre-training data. Specifically, we bridge the modality gap between a frozen image encoder and a large language model with a transformer and introduce two innovative learning strategies: noise-adaptive learning and concept-enhanced learning to mitigate the impact of noise. In noise-adaptive learning, we estimate the noise probability of each image-text pair based on the transformer's memorization effect and employ noise-adaptive regularization on image-text contrastive learning to condition cross-modal alignment. In concept-enhanced learning, we enrich incomplete text by incorporating visual concepts (objects in the image) to provide prior information about existing objects for image-text matching and image-grounded text generation, thereby mitigating text incompletion. Our framework effectively utilizes noisy web data and achieves state-of-the-art performance with less pre-training data across a wide range of vision-language tasks, including image-text retrieval, image captioning, and visual question answering.
\end{abstract}


\section{Introduction}

Vision-Language Models (VLMs) have recently achieved remarkable success across a variety of vision-language tasks\cite{zhou2024reconstruction,Li2024Vehicle,tao2023mlad,chen2024xmecap,chen2023mapo,kang2022tie,kang20216,zheng2024identification,zhang2024cu,zeng2024wordepth,chen2023xmqas,yang2024neurobind,chen2024hotvcom,yu2024advanced,yu2024credit,fan2024towards,fan2024advanced, tao2023sqba,sun2024enhancing,wei2024feature,xu2024investigating}. Central to these advancements is the pre-training of models on large-scale web-crawled datasets\cite{SBU2011Ordonez}\cite{Changpinyo2021Conceptual1P}\cite{YFCC100M}. These datasets provide a vast and diverse source of image-text pairs, which are crucial for training models that can generalize well to a wide range of downstream tasks. However, the noisy and incomplete nature\cite{tao2023mlad}\cite{tao2023sqba} of web data introduces significant challenges. The inherent inaccuracies and incompletion in these datasets necessitate scaling up the dataset size to achieve robust performance\cite{Qi2020ImageBERTCP}\cite{Pereyra2017RegularizingNN}\cite{Lukasik2020DoesLS}, making end-to-end training computationally prohibitive\cite{Chen2019UNITERUI}\cite{Wang2021SimVLMSV}.

Prior research has attempted to address this issue in various ways. Some methods employ filters to clean the data\cite{Sharma2018ConceptualCA}\cite{BLIP}\cite{jiaYXCPPLSLD21}, while others generate pseudo-targets to serve as auxiliary signals that help reduce the impact of noise\cite{Li2021AlignBF}\cite{Huang2021LearningWN}\cite{Reed2014TrainingDN}\cite{Arpit2017mem}. For example, models like ALBEF\cite{Li2021AlignBF} use momentum models to generate pseudo-targets, and BLIP\cite{BLIP} employs filters to remove noisy data and regenerate captions. However, these approaches often fail to simultaneously tackle both incorrect and incomplete data, limiting their effectiveness in fully mitigating the noise problem.

In this paper, we introduce NEVLP, an \textbf{N}oise-robust framework for \textbf{E}fficient \textbf{V}ision-\textbf{L}anguage \textbf{P}re-training, which designed to overcome the limitations of previous methods. Our approach leverages frozen pre-trained image encoders and large language models (LLM), bridging the modality gap with a lightweight transformer. We introduce two innovative learning strategies, noise-adaptive learning and concept-enhanced learning, to mitigate the impact of noise effectively. (1) Noise-Adaptive Learning: We estimate the noise probability of each image-text pair based on the transformer's memorization effect. This estimation allows us to apply noise-adaptive regularization on image-text contrastive learning, conditioning cross-modal alignment according to the noise probability and preventing the model from overfitting to noisy data. (2) Concept-Enhanced Learning: We enrich incomplete text by incorporating visual concepts (objects in the image) to provide prior information about existing objects in the image-text matching and image-grounded text generation. This strategy enhances the model's ability to generate more accurate and contextually relevant captions, thereby mitigating text incompletion. Our methodology offers several advantages over prior work. By employing frozen pre-trained models, we significantly reduce the computational cost associated with end-to-end training. Our noise-adaptive strategy effectively addresses both incorrect and incomplete data, enhancing the robustness of the pre-training process. The contributions of our paper are summarized as follows:
\begin{figure*}
    \centering
    \includegraphics[width=1\linewidth]{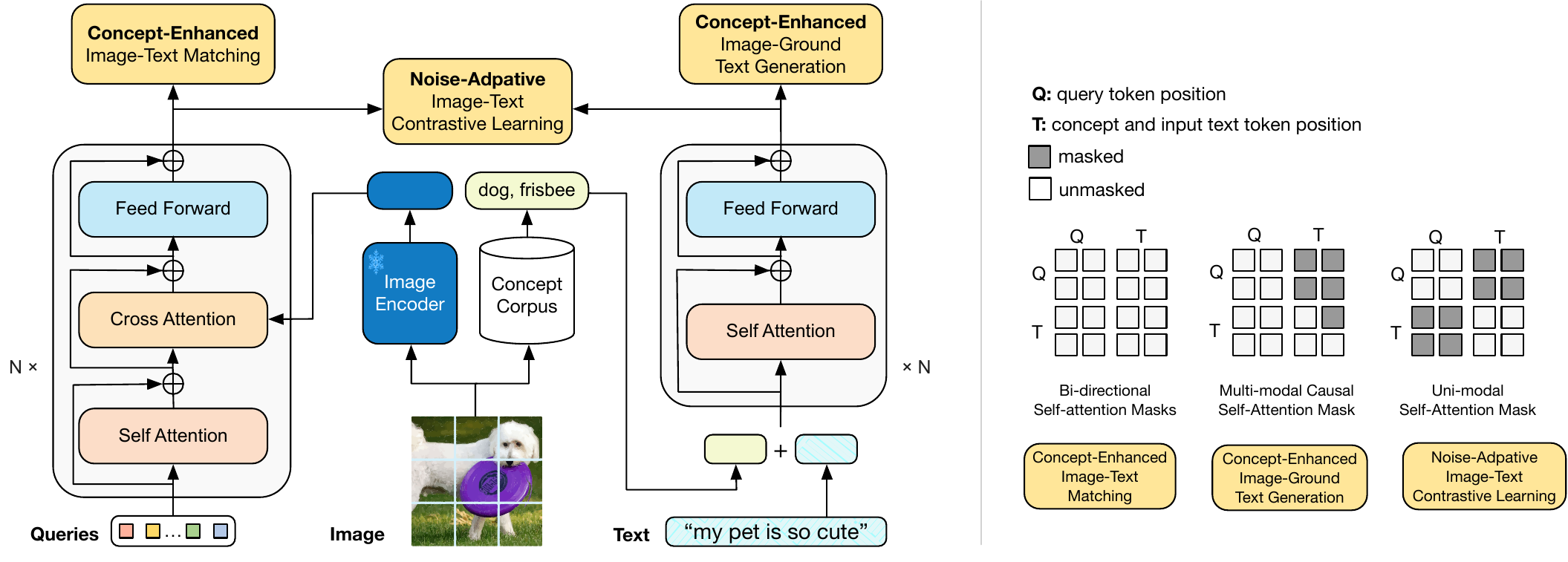}
    \caption{(\textbf{Left}) model architecture of NEVLP and the first stage vision language representation learning objectives. The framework consists of an image transformer for visual feature extraction and a text transformer that function as both a text encoder and a text decoder. Image transformer and text transformer share the same self-attention layer. We jointly optimize three objectives: concept-enhanced image-text matching, concept-enhanced image-ground text generation and noise-adpative image-text contrastive learning. (\textbf{Right}) The self-attention masking strategy for each objective to control query-text interaction.}
    \label{fig:model_framework}
\end{figure*}
\begin{itemize}
    \item We propose NEVLP, a noise-robust language-image pre-training framework that leverages frozen image encoders and LLMs. The NEVLP framework introduces a \textbf{noise-adaptive learning strategy} that effectively mitigate the impact of noisy data. Additionally, it incorporates \textbf{concept-enhanced learning strategy} to enrich incomplete text with visual concepts, improving the contextual relevance of generated captions.
    \item Our framework achieves state-of-the-art performance with less pre-training data across various vision-language tasks, including image-text retrieval, image captioning, and visual question answering.
\end{itemize}

\section{Method}
In this section, we describe NEVLP, our noise-robust vision-language pre-training framework. Inspired by BLIP2\cite{BLIP2}, we bridge the modality gap between the frozen image encoder and the frozen LLM using a transformer which is pre-trained in two stages. First, the transformer is pre-trained with the frozen image encoder to learn to extract text-relevant visual representations, incorporating noise-adaptive learning and concept-enhanced learning to handle noisy and incomplete samples. Second the transformer is pre-trained with the frozen LLM to utilize its generative language capabilities. We first present the model architecture and then describe two pre-training stages.
\subsection{Model Architecture}
Fig.\ref{fig:model_framework} illustrates the overall model architecture of NEVLP, which includes a concept corpus and two transformer modules sharing the same self-attention layer. The image transformer interacts with the frozen image encoder, using a set of learnable query embeddings as input visual feature extraction. Concept corpus contains visual concepts for each image and provides them as auxiliary inputs. The text transformer interacts with these visual concepts prepended to the text, functioning as both a text encoder and a text decoder.

\subsection{Vision-Language Representation Learning}
Vision-language representation learning aims to pre-train the transformer to extract the visual feature that are most relevant to the visual concept and text. We optimize three pre-training objectives and employ different attention masking strategies to control query-text interaction\cite{BLIP2}.

\textbf{Noise-adaptive Image-Text Contrastive Learning} (NITC) enhances the standard image-text contrastive (ITC) loss by incorporating the probability that an image-text pair does not semantically match. The ITC loss \cite{Li2021AlignBF} pulls the embeddings of positive image-text pairs $\{x_i, y_i\}$ closer while pushes those of negative pairs $\{x_i, y_j\}_{i \ne j}$ apart. Let $\mathcal{V}(x_i)$ and $\mathcal{T}(y_j)$ represent the normalized features from image and text transformers, respectively. Both transformers share the same self-attention layer, and a unimodal self-attention are applied to ensures that queries and texts do not interact. The pre-sample similarity between image-text pairs is given by $s^{y}_{i,j}=s^{x}_{i,j}=\mathcal{V}(x_i) ^{\top}\mathcal{T}(y_j)$. The ITC loss maximizes similarity for positive pairs and minimizes it for negative pairs. NITC improves upon this by factoring in the noisy probability that a positive pair might not be semantically aligned. Inspired by NLIP\cite{NLIP_2023}, a two component Gaussian Mixture Model\cite{Permuter2006ASO} is used to fit the pre-sample ITC loss, with the higher mean component predicting the noisy probability $\varepsilon_{i}$ for the $i$-th image-text pair. Given a batch size $B$ and a smoothing rate $\omega_i = \lambda \varepsilon_i$ with hype-parameter $\lambda$, NITC incorporates noise-adaptive regulation to prevent overfitting on noisy pairs as follows: 
\begin{equation}
   \mathcal{L_{\mathrm{NITC} }} = \frac{1}{2B}\sum_{i=1}^{B}(\mathcal{L}^{x}_i + \mathcal{L}^{y}_i)  
\end{equation}
\begin{equation}
   \mathcal{L} _{i}^{x} = -\mathrm{log}\frac{(1-\omega_i)\mathrm{exp} (s_{i,i}^{x}) }{(1-\omega_i)\mathrm{exp} (s_{i,i}^{x}) + \frac{\omega_i}{B-1}\sum_{i\ne j}  \mathrm{exp} (s_{i,j}^{x})  } 
\end{equation}
\begin{equation}
   \mathcal{L} _{i}^{y} = -\mathrm{log}\frac{(1-\omega_i)\mathrm{exp} (s_{i,i}^{y}) }{(1-\omega_i)\mathrm{exp} (s_{i,i}^{y}) + \frac{\omega_i}{B-1}\sum_{i\ne j}  \mathrm{exp} (s_{i,j}^{y})  } 
\end{equation}

\textbf{Concept-Enhanced Image-Ground Text Generation} (CITG) loss trains a text transformer to generate descriptions that combine both visual features from the image transformer and additional visual concepts from a concept corpus. Web-crawled image-text data often lacks detail, with the text frequently missing key visual elements from the image\cite{BLIP}. CITG addresses this issue by enriching incomplete text with visual concepts to provide additional context. Following NLIP\cite{NLIP_2023}, we create a large visual concept corpus $Q$ by extracting concept nouns from a web-collected corpus. For each image, we retrieve the top-k words as visual concept $q_i \in Q$ based on image-text similarity $sim(x_i,Q)$. This process uses a pre-trained VLM with $\mathcal{V}_p$ as the image encoder and $\mathcal{T}_p$ as the text encoder. To bridge the gap with natural language, a pre-defined text prompt $p$ is combined with the visual concepts\cite{Radford2021LearningTV}. The similarity between image $x_i$ and nouns in Q is defined as:
\begin{equation}
    sim(x_i,Q) = \left \langle \mathcal{V}_p(x) \cdot \mathcal{T}_p([p,Q])  \right \rangle 
\end{equation}
We use a multimodal causal self-attention mask to control query-text interaction \cite{unilm2019}. Queries can attend to each other but not to visual concepts or text tokens. Text tokens can attend to all queries, visual concepts, and preceding text tokens. A \texttt{[DEC]} token is used as the initial token to indicate the start of the text generation task. The task is then optimized using language modeling loss.

\textbf{Concept-Enhanced Image-Text Matching} (CITM) is a binary classification task that predicts whether an image-text pair is matched (positive) or unmatched (negative). Visual concepts are prepended to the input text to provide additional context for cross-modal matching. A bi-directional self-attention mask allows queries, visual concepts, and text to attend to each other. The output query embeddings are fed into a linear classifier, and the average logits across all queries are used as the matching score. A hard negative mining strategy \cite{BLIP} is employed to generate informative negative pairs.

\subsection{Vision Language Generative Learning}
In vision-language generative learning, the transformer is pre-trained with the frozen LLM to leverage the LLM’s language generation capabilities. The image transformer's output query embeddings are projected into the LLM's text embedding space through a fully-connected (FC) layer. During the vision-language representation learning stage, the transformer extracts highly informative visual representations in the query embeddings, employing noise-adaptive learning and concept-enhanced learning to address noise issues. It reduces the LLM’s workload in understanding the image and generating related text. In the vision language generative learning stage, a decoder-based LLM generates text from the query embeddings, and language modeling loss is used to further pre-train both the transformer and the FC layer.

\section{Experiment}
\subsection{Experimental Settings}
\textbf{Forzen image encoder and LLM.} We use two frozen image encoders: (1) ViT-L/14 from CLIP \cite{Radford2021LearningTV} and (2) ViT-g/14 from EVA-CLIP \cite{EVA-CLIP}. We use the output features from the second-to-last layer of ViT. For the frozen LLMs, we adopt Llama2 \cite{llama2023touvron} with parameters of 7B, 13B, and 70B as our decoder-based LLMs.

\textbf{Visual Concept Corpus.} We build the visual concept corpus following the approach in NLIP\cite{NLIP_2023} by parsing nouns from the source corpus. The source corpus includes YFCC100M\cite{YFCC100M}, WordNet of Natural Language Toolkit\cite{loper-bird-2002-nltk}, OpenWebText\cite{Gokaslan2019OpenWeb}. Using the SpaCy toolkit, we extract nouns, filter out those appearing fewer than five times, and compile a final visual concept corpus containing 151k unique nouns. We use a pre-trained $\mathrm{FILIP_{large}}$ \cite{Yao2021FILIPFI} to retrieve nouns with top-3 image-text similarity for each image as the visual concepts.

\textbf{Pre-training.} Our pre-training dataset consists of a total of 42M images, including two human-annotated datasets (COCO \cite{MicrosoftCC2014Lin} and Visual Genome \cite{visualGenome2017}) and four web datasets (CC3M \cite{Changpinyo2021Conceptual1P}, CC12M \cite{Changpinyo2021Conceptual1P}, SBU \cite{SBU2011Ordonez}, and YFCC100M \cite{YFCC100M}). We apply filtering rules based on FILIP \cite{Yao2021FILIPFI}. Images are randomly cropped and resize to 224 $\times$ 224 during pre-training, with resolution increased to 384 $\times$ 384 for downstream tasks. We use the AdamW optimizer with a weight decay of 0.05, a peak learning rate of 1e-4, and cosine learning rate decay. The batch size is 1600 during vision-language representation learning and 1440 during vision-language generative learning. We follow \cite{li2019segmentation} to tackle the class imbalance problem. In the vision-language representation learning phase, we first warm up the NEVLP with ITC, CITG, and CITM losses for 1k epochs. We then estimate the noisy probability $\varepsilon_i$ for each image-text pair based on ITC loss and apply noise-adaptive regularization for the subsequent 100k epochs. To obtain high-quality captions, we fine-tune the text transformer on COCO Captions \cite{MicrosoftCC2014Lin}, generate concept-enhanced text, and replace the old text for each image-text pair. Finally, we further pre-train the NEVLP on the revised image-text pairs for 100k epochs. In the vision-language generative learning phase, the transformer and FC layer are connected to the frozen LLM and trained with revised image-text pairs for 50k epochs.

\begin{table}[htbp]
\caption{Comparision with state-of-the-art image captioning methods.}
\begin{tabular}{c|c|c|c}
\hline
Model                 & Pre-train images & BLEU@4          & CIDEr \\ \hline
SimVLM\cite{Wang2021SimVLMSV}                & 1.8B             & 39.0            & 134.8 \\
Flamingo\cite{Flamingo2024}              & 1.8B             & -               & 138.1 \\
BLIP\cite{BLIP}                  & 129M             & 39.7            & 133.3 \\
BLIP2\cite{BLIP2}                 & 129M             & 42.4            & 144.5 \\ \hline
NEVLP ViT-g Llama2 7B  & 42M              & 40.7            & 141.8 \\
NEVLP ViT-g Llama2 13B & 42M              & 42.5            & 143.4 \\
NEVLP ViT-g Llama2 70B & 42M              & \textbf{44.2}   & \textbf{146.4} \\ \hline
\end{tabular}
\label{table:image_caption}
\end{table}

\subsection{Image Captioning}
We evaluate our NEVLP model on the image captioning task, which involves generating a textual description for a given image. We fine-tune the NEVLP model on the COCO dataset using language modeling loss and assess its performance on the COCO test set with standard metrics, including BLEU and CIDEr. As shown in Table \ref{table:image_caption}, NEVLP achieves the best performance even with a smaller scale of pre-trained images, scoring 44.2 in BLEU@4 and 146.4 in CIDEr. This performance surpasses BLIP2 \cite{BLIP2} by 1.8 in BLEU@4 and 1.9 in CIDEr.  NEVLP also beat other models which are pre-trained on large scale datasets.

\subsection{Image-Text Retrieval}
In the image-text retrieval task, the goal is to find the most relevant images for a given textual query or the most relevant text descriptions for a given image query. We evaluate the NEVLP model on both image-to-text and text-to-image retrieval tasks. Since this task does not involve language generation, the frozen LLM is not involved. Instead, we fine-tune the image encoder and transformer on COCO dataset using NITC, CITM, and CITG losses. The experiments use ViT-L and ViT-g as the image encoders, and evaluation is performed on the Flickr30K \cite{Flickr30k2015} dataset. Following \cite{Li2021AlignBF}, we first select k=128 candidates based on image-text feature similarity and then rerank them based on pairwise CITM scores.  
\begin{table}[htbp]
\caption{Image to Text and Text to Image Retrieval on Flickr30K.}
\scriptsize
\begin{tabular}{c|c|cccccc}
\hline
\multirow{3}{*}{Model} & \multirow{3}{*}{\begin{tabular}[c]{@{}c@{}}Pre-train\\ images\end{tabular}} & \multicolumn{6}{c}{Flickr30K (1K test set)}                                                      \\
                       &                                                                             & \multicolumn{3}{c}{Image-To-Text}                & \multicolumn{3}{c}{Text-To-Image}             \\
                       &                                                                             & R@1           & R@5  & R@10                      & R@1           & R@5           & R@10          \\ \hline
CLIP\cite{Radford2021LearningTV}                   & 400M                                                                        & 88.0          & 98.7 & \multicolumn{1}{c|}{99.4} & 68.7          & 90.6          & 95.2          \\
FILIP\cite{Yao2021FILIPFI}                  & 400M                                                                        & 89.8          & 99.2 & \multicolumn{1}{c|}{99.8} & 75.0          & 93.4          & 96.3          \\
ALBEF\cite{Li2021AlignBF}                  & 14.1M                                                                       & 94.1          & 99.5 & \multicolumn{1}{c|}{99.7} & 82.8          & 96.3          & 98.1          \\
BLIP\cite{BLIP}                   & 129M                                                                        & 96.7          & 100  & \multicolumn{1}{c|}{100}  & 86.7          & 97.3          & 98.7          \\
BLIP2\cite{BLIP2}                  & 129M                                                                        & 96.9          & 100  & \multicolumn{1}{c|}{100}  & 88.6          & 97.6          & 98.9          \\ \hline
NEVLP ViT-L            & 42M                                                                         & 96.5          & 100  & \multicolumn{1}{c|}{100}  & 89.7          & 98.1          & 98.9          \\
NEVLP ViT-g            & 42M                                                                         & \textbf{97.1} & 100  & \multicolumn{1}{c|}{100}  & \textbf{89.9} & \textbf{98.4} & \textbf{98.9} \\ \hline
\end{tabular}
\label{table:image_text_retrieval}
\end{table}

As shown in Table \ref{table:image_text_retrieval}, NEVLP achieves the best performance, with significant improvement in both image-to-text retrieval and text-to-image retrieval compared to the state-of-the-art models. NEVLP slightly outperforms BLIP2 by +0.2\% in average recall@1 for image to text retreival and shows improvements of 1.3\% in average recall@1 and 0.8\% in average recall@5 for text-to-image retrieval, achieving the same performance in average recall@10 as BLIP2. Note that BLIP2 is pre-trained with 3$\times$ image-text pairs than NEVLP.

\subsection{Visual Question Answering}
Visual Question Answering (VQA) involves answering questions based on a given image. It requires the model to understand both the visual content of the image and the semantics of the question to generate accurate answers. Following \cite{Li2021AlignBF}, we treat VQA as an answer generation task and use an open-ended answer generation loss to fine-tune the transformer and image encoder while keeping the LLM frozen. The frozen LLM receives the projected output query embeddings from the transformer and the question as input to generate the answer. Following BLIP2 \cite{BLIP2}, the question token is input into the transformer's self-attention layer to help the model focus on question-related image regions. Fine-tuning and evaluation are performed on the VQAv2 \cite{VQAv2} dataset. The results are shown in Table \ref{table:vqa}. Using only 42M pre-train images, NEVLP outperforms BLIP2 by +1.96\% on development test set (test-dev) and +1.99\% on standard test set (test-std). NEVLP also outperforms other existing methods by a significant margin.

\begin{table}[htbp]
\caption{Comparison with state-of-art methods on VQAv2.}
\begin{tabular}{c|c|cc}
\hline
\multirow{2}{*}{Model} & \multirow{2}{*}{Pre-train Images} & \multicolumn{2}{c}{VQAv2}       \\
                       &                                   & test-dev       & test-std       \\ \hline
ALBEF\cite{Li2021AlignBF}                  & 14.1M                             & 75.84          & 76.04          \\
Flamingo\cite{Flamingo2024}               & 1.8B                              & 82.00          & 82.10          \\
BLIP\cite{BLIP}                   & 129M                              & 78.25          & 78.32          \\
BLIP2\cite{BLIP2}                  & 129M                              & 81.55          & 81.66          \\ \hline
NEVLP ViT-g  Llama2 7B & 42M                               & 82.57          & 82.68          \\
NEVLP ViT-g Llama2 13B & 42M                               & 83.36          & 83.48          \\
NEVLP ViT-g Llama2 70B & 42M                               & \textbf{83.51} & \textbf{83.65} \\ \hline
\end{tabular}
\label{table:vqa}
\end{table}

\begin{table}[htbp]
\caption{Ablation studies on 3 downstream tasks.}
\begin{tabular}{c|cc|cc|cc}
\hline
\multirow{2}{*}{} & \multicolumn{2}{c|}{Caption(CoCo)} & \multicolumn{2}{c|}{Retrival(Flickr)} & \multicolumn{2}{c}{VQA(VQAv2)}  \\
                  & BLEU@4          & CIDEr            & TR@1              & IR@1              & test-dev       & test-std       \\ \hline
NEVLP             & \textbf{40.7}   & \textbf{141.8}   & \textbf{97.1}     & \textbf{89.9}     & \textbf{82.57} & \textbf{82.68} \\
w/o NA            & 39.8            & 139.1            & 93.1              & 84.8              & 79.71          & 79.93          \\
w/o CE            & 39.3            & 137.5            & 94.9              & 86.4              & 80.12          & 80.45          \\ \hline
\end{tabular}
\label{table:ablation}
\end{table}
\subsection{Ablation Studies}
Table \ref{table:ablation} presents the results of ablation studies on the NEVLP ViT-g Llama2 7B model across downstream tasks, including image captioning, image-text retrieval, and VQA. We denote the noise-adaptive strategy as NA and the concept-enhanced strategy as CE. For simplicity, recall@1 for image-to-text retrieval and text-to-image retrieval are denoted as TR@1 and IR@1, respectively.

\textbf{Effect of Noise-adaptive Learning.} Table \ref{table:ablation} shows that the noise-adaptive learning improves the model's performance on all downstream tasks. The most notable gains are observed in image-text retrieval, with a +4.0\% boost in TR@1 and a +5.1\% boost in IR@1. For VQA, the noise-adaptive strategy enhances performance by +2.86\% on test-dev and +2.75\% on test-std. It also slightly improves performance on image captioning, with a +0.9\% gain in BLEU@4 and +2.7\% in CIDEr. These results demonstrate that noise-adaptive learning helps avoid model overfitting on semantically mismatched pairs.

\textbf{Effect of Concept-Enhanced Learning.} Table \ref{table:ablation} indicates that the concept-enhanced learning strategy improves image captioning performance by 1.4\% in BLEU@4 and 4.3\% in CIDEr, confirming that pre-training with CITM and CITG losses helps capture more information about existing objects and enriches the generated captions. Without concept-enhanced learning, performance on other downstream tasks also declines significantly.

\section{Conclusion}
The paper proposes NEVLP, a noise-robust VLP framework for efficient vision-language pre-training. NEVLP employs two learning strategy to effectively utilize web-crawled datasets. Noise-adaptive learning prevents overfitting on sematically mismatched image-text pairs by applying noise-adaptive regularization during image-text contrastive learning. Concept-enhanced learning addresses the issue of incomplete web text by incorporting visual concepts into image-text matching and image-grounded text generation. Experimental results demonstrate that NEVLP achieves state-of-the-art performance on various vision-language tasks while requiring less pre-training data. In the future, we plan to integrate these two learning strategies into more cross-modal pre-training models that require noise-robust capabilities.

\bibliographystyle{IEEEtran}
\bibliography{IEEEabrv, ref.bib}

\vspace{12pt}

\end{document}